\documentclass[a4paper,fleqn]{cas-dc}
\usepackage{placeins}
\usepackage[square, comma, sort&compress, numbers]{natbib}
\usepackage{rotating}
\usepackage{url}
\usepackage{bm}
\usepackage{algorithm}
\usepackage[noend]{algpseudocode}
\usepackage{caption} 
\usepackage{setspace}
\usepackage{tabularx} 
\usepackage{booktabs} 
\usepackage{times}     
\usepackage{xurl}   
\usepackage{caption}
\usepackage{booktabs}   
\usepackage{ulem}       
\usepackage[utf8]{inputenc}

\begin{document}
\let\WriteBookmarks\relax
\def\floatpagepagefraction{1}
\def\textpagefraction{.001}

\shorttitle{Twisted Convolutional Networks (TCNs): Enhancing Feature Interactions for Non-Spatial Data Classification}

\shortauthors{Lian J.J. et al.}

\title[mode=title]{Twisted Convolutional Networks (TCNs): Enhancing Feature Interactions for Non-Spatial Data Classification}            

\author[1]{Junbo Jacob Lian}[orcid=0000-0001-7602-0022]
\ead{jacoblian@u.northwestern.edu}
\credit{Formal Analysis, Investigation, Resources, Conceptualization, Methodology, Software, Data Curation, Visualization, Writing – original draft, Writing – review \& editing.}
\affiliation[1]{organization={McCormick School of Engineering, Northwestern University},%
               city={Evanston, IL},%
               country={USA}}

\author[2]{Haoran Chen}[orcid=0009-0001-2415-0033]
\ead{chr@stu.zafu.edu.cn}
\credit{Formal Analysis, Investigation, Data Curation, Validation.}
\affiliation[2]{organization={School of Mathematics and Computer Science, Zhejiang A\&F University},%
               city={Hangzhou},%
               country={PR China}}
               
\author[3]{Kaichen Ouyang}[orcid=0009-0003-5937-5229]
\ead{oykc@mail.ustc.edu.cn}
\credit{Formal Analysis, Investigation, Resources, Software.}
\affiliation[3]{organization={School of Mathematics, University of Science and Technology of China},%
               city={Hefei},%
               country={PR China}}

\author[4]{Yujun Zhang}[orcid=0000-0003-3016-8843]
\ead{zhangyj069@gmail.com}
\credit{Validation, Writing – review \& editing.}
\affiliation[4]{organization={College of New Energy, Jingchu University of Technology},%
               city={Jingmen},%
               country={PR China}}

\author[5]{Rui Zhong}[orcid=0000-0003-4605-5579]
\ead{zhongrui@iic.hokudai.ac.jp}
\credit{Validation, Writing – review \& editing.}
\affiliation[5]{organization={Information Initiative Center, Hokkaido University},%
               city={Sapporo},%
               country={Japan}}

\author[6]{Huiling Chen}[orcid=0000-0002-7714-9693]
\cormark[1]
\ead{chenhuiling.jlu@gmail.com}
\credit{Formal Analysis, Funding Acquisition, Supervision, Writing – review \& editing.}
\affiliation[6]{organization={School of Computer Science and Artificial Intelligence, Wenzhou University},%
               city={Wenzhou},%
               country={PR China}}
               
\cortext[1]{Corresponding author.}

\begin{abstract}
Twisted Convolutional Networks (TCNs) are proposed as a novel deep learning architecture for classifying one-dimensional data with arbitrary feature order and minimal spatial relationships. Unlike conventional Convolutional Neural Networks (CNNs) that rely on structured feature sequences, TCNs explicitly combine subsets of input features through theoretically grounded multiplicative and pairwise interaction mechanisms to create enriched representations. This feature combination strategy, formalized through polynomial feature expansions, captures high-order feature interactions that traditional convolutional approaches miss. We provide a comprehensive mathematical framework for TCNs, demonstrating how the twisted convolution operation generalizes standard convolutions while maintaining computational tractability. Through extensive experiments on five benchmark datasets from diverse domains (medical diagnostics, political science, synthetic data, chemometrics, and healthcare), we show that TCNs achieve statistically significant improvements over CNNs, Residual Networks (ResNet), Graph Neural Networks (GNNs), DeepSets, and Support Vector Machine (SVM). The performance gains are validated through statistical testing. TCNs also exhibit superior training stability and generalization capabilities, highlighting their robustness for non-spatial data classification tasks. Source code is publicly available at \url{https://github.com/junbolian/Twisted-Convolutional-Networks}.
\end{abstract}

\begin{keywords}
Neural networks \sep Machine learning \sep Feature combination \sep Twisted Convolutional Networks \sep Non‑spatial data \sep Polynomial feature expansion
\end{keywords}

\maketitle

\section{Introduction} \label{sec:1}

Recent advancements in machine learning and deep learning have revolutionized classification and pattern recognition. Convolutional Neural Networks (CNNs) in particular have achieved remarkable success by capturing spatial hierarchies in data \cite{wang2024blockchain, rubaiyat2024signal, zhang2023ecg}, making them highly effective for tasks such as image and speech recognition \cite{lecun2015deep, liang2024crop, ma2023hyperspectral, fan2025hcpa}. Despite their success in domains where feature locality and order are informative, CNNs heavily rely on the spatial or sequential order of input features. This reliance can limit their applicability to data without an inherent spatial structure or well-defined feature relationships. In many real-world applications—such as gene expression profiles, customer demographic data, or multi-sensor readings—the relationships between features are not spatially local or sequential, and the ordering of features may carry little meaningful information \cite{larranaga2006mlbio, zaharia2012resilient, domingos2012useful}. In these cases, CNNs often fail to achieve optimal performance \cite{li2021survey} because their convolutional filters cannot effectively capture the complex, global interactions among unordered features \cite{bello2025geconv}.

To address these limitations, researchers have explored alternative architectures that better handle unordered or independent features. For instance, attention mechanisms \cite{vaswani2017attention} can capture pairwise relationships by weighting feature importance dynamically, graph neural networks (GNNs) \cite{kipf2017semi} model data as arbitrary graphs of inter-feature connections, and permutation-invariant networks like DeepSets \cite{zaheer2017deep} ensure that model outputs do not depend on input feature ordering. These approaches have seen success in domains where feature order is irrelevant, such as point cloud classification \cite{qi2017pointnet} and set anomaly detection \cite{mohammad2023one}. However, many of these methods still do not fully exploit the information embedded in explicit combinations of features. They either aggregate features without modeling interactions (simple pooling in DeepSets) or require known relational structure ( adjacency in GNNs), or they implicitly consider all interactions without learning which are most useful (as in attention mechanisms that attend to every pair).

Kernel methods, such as Support Vector Machines (SVMs) with non-linear kernels, provide another way to capture complex feature interactions. Radial Basis Function (RBF) kernels and polynomial kernels implicitly consider high-order combinations of input features \cite{scholkopf2002learning}. In theory, an SVM with a sufficiently rich kernel (like an RBF kernel) can approximate any decision boundary by mapping data into a high-dimensional feature space. In practice, however, kernel methods often suffer from high computational cost and poor scalability to large datasets, as well as difficulties in interpreting or selecting the most relevant feature interactions \cite{lian2025prickly}. For example, an SVM with a polynomial kernel of degree $d$ expands feature interactions up to $d$-way products, but it treats all such interactions uniformly and the number of terms grows combinatorially, making it hard to focus on the most informative combinations \cite{burges1998tutorial}.

In this paper, we propose the \textbf{Twisted Convolutional Networks (TCNs)} to overcome the challenges posed by unordered features and to explicitly model complex feature interactions. TCN introduces a novel operation called \emph{twisted convolution}, which generalizes the idea of convolution to operate on combinations of features rather than on spatially neighboring features. Instead of using fixed contiguous kernels as in CNNs, a twisted convolution layer generates new feature maps by applying learned operations on various subsets of the input features. This approach is inspired by the idea of polynomial feature expansion from kernel methods, but implemented in a trainable, data-driven manner. By systematically combining features (through both multiplicative and additive interactions), TCN is able to capture higher-order relationships in the data while remaining agnostic to the input order of features. Essentially, TCN learns which feature interactions are important for the task, rather than assuming a particular structure a priori.

We hypothesize and show empirically that this strategy yields richer and more informative representations, leading to improved performance on classification tasks where features have no natural ordering or local correlation. We validate the effectiveness of TCN across five benchmark datasets and compare its performance to traditional CNNs and other state-of-the-art models (ResNet, GNN, DeepSets), as well as a strong classical baseline (SVM with RBF kernel). Our results demonstrate that TCN not only achieves the highest accuracy in most cases, but also maintains better training stability and generalization, suggesting a lower risk of overfitting on these tasks. 

The key contributions of this paper are as follows:

\begin{itemize}
    \item \textbf{A Novel Twisted Convolution Operation:} We introduce TCN as a new deep network architecture with a mathematically-grounded twisted convolution layer. This operation explicitly combines input features in multiplicative and pairwise ways, effectively generalizing standard convolution to non-spatial feature spaces. We provide a formal definition of twisted convolution and show how it can be seen as an explicit polynomial expansion with learnable coefficients, giving TCN the power to capture complex interactions among features.
    \item \textbf{Enhanced Feature Interaction Modeling:} The TCN architecture is designed to capture high-order feature interactions that other models might miss. By generating new feature representations from subsets of the original features, TCN uncovers patterns that are not apparent when considering features individually or only in fixed local groups. We discuss how our approach differs from and improves upon existing methods (CNNs, GNNs, etc.), and we analyze the impact of our feature combination strategies on mitigating the dependency on feature ordering.
    \item \textbf{Comprehensive Empirical Evaluation:} We conduct extensive experiments on five diverse datasets and provide a thorough comparison between TCN and several baseline models: CNN, ResNet, GNN, DeepSets, and SVM. Our evaluation spans multiple domains to demonstrate the generality of TCN. We report not only accuracy but also precision, recall, and F1-score for a holistic assessment. The results show that TCN achieves superior performance across most metrics and datasets. We also observe that TCN offers more stable training dynamics (smaller fluctuations in accuracy) and better generalization (smaller gap between training and test performance) compared to the baselines.
    \item \textbf{Analysis of Model Behavior:} Beyond raw performance, we analyze how the number of feature combinations ($C$) in TCN affects its learning capacity and risk of overfitting. By varying $C$ (considering pairs, triples, or quadruples of features in combinations) and evaluating on different datasets, we provide insights into selecting this hyperparameter for optimal performance. This analysis highlights the trade-off between model expressiveness and generalization in the context of feature interaction modeling.
\end{itemize}

The remainder of this paper is organized as follows: Section \ref{sec:2} reviews related work, covering prior approaches to learning from unordered features and capturing feature interactions. Section \ref{sec:3} presents the Twisted Convolutional Network in detail, including the theoretical formulation of twisted convolution, the network architecture, and training considerations. Section \ref{sec:4} describes the baseline models and the experimental setup for our evaluation. Section \ref{sec:5} reports the experimental results, comparing TCN with other methods and discussing the findings, including ablation on feature combination strategies. Finally, Section \ref{sec:6} concludes the paper and suggests directions for future work.

\section{Related Work}\label{sec:2}

Learning from data with unordered or non-spatial features has been approached from various angles in the literature. Here, we discuss several threads of related work, including adaptations of CNNs for non-spatial data, kernel methods for capturing feature interactions, attention mechanisms, graph-based methods, ensemble learning and feature engineering techniques, as well as specialized networks for set inputs and explicit feature interaction modeling \cite{saha2024groupfeature}.

\subsection{CNNs and Adaptations for Non-Spatial Data}
CNNs are best known for their success in domains like image and audio processing, where they exploit local spatial or temporal structure via convolutional filters \cite{krizhevsky2012imagenet, li2025cnn-informer, girshick2015fast, hinton2012deep}. In scenarios where such structure is absent, a standard CNN’s performance can degrade. Researchers have attempted to adapt CNNs to non-spatial data by treating features as a one-dimensional sequence (in an arbitrary or given order) and applying 1D convolutions. Lee \cite{lee2007nonlinear} explored using CNN-like architectures on non-grid data by imposing an order or using fully connected layers to mimic convolutional feature extractors. However, these adaptations still impose a form of locality or sequence on the features, which might not align with the underlying data characteristics \cite{ali2022predictive, heaney2024applying, chen2025intra}. In contrast, our TCN does not assume any fixed ordering or locality; instead, it learns feature groupings based on predictive value.

\subsection{Kernel Methods for Feature Interactions}
Kernel methods, particularly Support Vector Machines with non-linear kernels, have long been used to capture interactions between features in a high-dimensional space \cite{scholkopf2002learning}. The polynomial kernel, for instance, expands the feature space to include all products of $d$ features (for a degree-$d$ polynomial), thereby modeling interactions up to that order. An RBF kernel, on the other hand, implicitly considers infinite-degree interactions by measuring distances in the original space. While powerful, these kernel approaches face challenges: they typically result in models with high computational complexity in both training and prediction (especially for large datasets, since kernel methods often require computing and storing an $N \times N$ Gram matrix for $N$ samples). Furthermore, kernels treat feature interactions implicitly and uniformly—there is no mechanism to learn which specific interactions are most important. In contrast, TCN provides an explicit and trainable approach: it constructs feature interaction terms (analogous to polynomial terms) in a neural network framework, with learnable weights that can zero in on the most informative combinations and ignore spurious ones. Connections between neural networks and kernel methods have been explored (deep networks as learned kernels \cite{cho2009kernel} or using random Fourier features to approximate kernels \cite{rahimi2007random}), but our work integrates the idea of explicit feature combination into the architecture of the network itself.

\subsection{Attention Mechanisms and Transformers}
Attention mechanisms \cite{vaswani2017attention, bahdanau2015neural} allow a model to weigh the importance of different input components dynamically. In Transformer models and related architectures, self-attention can relate any pair of positions in the input by computing similarity scores and attending accordingly. Vision Transformers (ViT) \cite{dosovitskiy2021image} demonstrated that even image data (traditionally the domain of CNNs) can be effectively processed by treating an image as a sequence of patches and applying self-attention, thus capturing long-range dependencies \cite{lian2025handbook}. For non-spatial tabular data, one could apply a self-attention mechanism to the feature vector, allowing the model to learn pairwise relationships between features. While this can capture interaction to an extent, attention typically combines features in a linear weighted sum fashion rather than forming new feature products or nonlinear combinations. Moreover, attention has quadratic complexity in the number of features, which could be inefficient when the feature dimensionality is high. TCN’s twisted convolution differs by explicitly creating new features from combinations, rather than reweighting existing features. Because Transformer architectures scale with $O(n^2)$ complexity and large parameter counts, they are not well matched to the small, low\textendash dimensional datasets considered in this study; accordingly, we omit Transformer\textendash style baselines and concentrate our comparisons on more computationally comparable models (Section~\ref{sec:4}).

\subsection{Graph Neural Networks}
Graph Neural Networks (GNNs) \cite{kipf2017semi, hamilton2017inductive, wu2021self} are a class of deep learning models that generalize neural networks to graph-structured data. If one interprets each feature as a node in a graph and draws edges to represent relationships (such as statistical correlations or domain-known associations), a GNN can propagate information between features according to this graph. Graph Convolutional Networks (GCNs) and Graph Attention Networks (GATs) \cite{velivckovic2018graph} could, in principle, learn feature interactions by treating the feature set as a fully connected graph (where every feature is connected to every other). Some recent works (in social science or biology) have constructed graphs of features to apply GNNs for feature selection or importance estimation \cite{tian2023asagnn, hu2023costgnn, lee2024analysis}. However, using a fully connected feature graph quickly becomes intractable as the number of edges grows quadratically with number of features, and it still requires either assuming some structure or learning an adjacency matrix \cite{fan2025path-aware, ye2025hyperbolic}. TCN avoids this by directly constructing combined features without needing an explicit graph of feature connections.

\subsection{Ensemble Learning and Feature Engineering}
Ensemble methods like Random Forests \cite{breiman2001random} implicitly consider multiple feature combinations by training many decision trees on random subsets of features. Each tree might capture different interactions, and the forest aggregates them. This is a form of explicit feature combination in a broader sense—different trees explore different subsets of features (like random $k$-way combinations) to make decisions. The success of Random Forests and related techniques (Gradient Boosted Trees) in tabular data competitions underlines the importance of modeling feature interactions and not relying on any fixed feature ordering \cite{rabbani2025deep, couplet2024investigating}. Similarly, manual feature engineering often involves creating new features by combining existing ones (products, ratios, differences, logical combinations, etc.), which can dramatically improve model performance \cite{guyon2003introduction, koza1994genetic}. Our TCN takes inspiration from these practices but automates them: it performs a learned combination of features analogous to an automated feature engineer within a neural network.

Deep Forest \cite{zhou2019deepforest} is another approach combining ensemble ideas with deep architectures, using cascades of random forests to progressively transform features. While effective, it still relies on decision tree logic rather than continuous optimization as in neural networks, and it doesn't explicitly form multiplicative combinations of features as TCN does.

\subsection{Permutation-Invariant Neural Networks}
DeepSets \cite{zaheer2017deep} is a seminal work on networks for set inputs (which are unordered by nature). A DeepSets model computes representations for each element (feature) via some function $\phi(x_i)$, then aggregates these (by summation or averaging) and processes the aggregate through another function $\rho$ to produce an output. This ensures the output is invariant to the input order. For sets of features (as opposed to sets of data points), one could treat each feature’s value as an “element” of the set and apply a similar approach. However, DeepSets in its basic form does not explicitly account for feature-feature interactions beyond what might be captured in the $\rho$ function after summing. It's effectively modeling a sort of global average of individual feature contributions, which might miss conjunctive conditions (cases where feature $A$ is only important in the presence of feature $B$). Extensions like Set Transformers introduce attention among elements to capture some interactions, but again in a soft additive way rather than creating new multiplicative features.

PointNet \cite{qi2017pointnet} and its extension PointNet++ \cite{qi2017pointnetpp} also operate on sets (specifically, point clouds). PointNet uses a similar approach to DeepSets (per-point processing and global pooling), while PointNet++ introduces hierarchical grouping of points to capture local structures. Analogously, one could hierarchically cluster features and aggregate within clusters, but defining such clusters for general data is non-trivial.

\subsection{Explicit Feature Interaction Models}
Beyond our work, there have been some neural network architectures aimed at modeling feature interactions explicitly, especially for high-dimensional tabular data such as those found in online advertising or recommender systems. For example, Deep Cross Networks (DCN) \cite{wang2021dcn} include cross layers that multiply features from previous layers with the original input features, thereby explicitly creating cross terms. TabNet \cite{arik2021tabnet} employs attentive feature selection at multiple decision steps, effectively learning which subset of features to consider together at each stage. Although not forming products, this mechanism can learn to focus on interactions by selecting groups of features. Feature-wise linear modulation (FiLM) \cite{perez2018film} can also condition one set of features on another through learned scaling, again an implicit interaction modeling.

Our TCN differs from these in that it directly constructs new features by combining existing ones through learned operations (multiplication or addition of products), within a convolution-like framework that is integrated into the network’s layers. It provides a more general and systematic way to generate high-order features, as opposed to adding a limited number of cross terms or gating features in/out.

In summary, the landscape of methods for handling unordered features and capturing feature interactions includes kernel methods, attention-based models, GNNs, ensembles, and specialized deep networks. TCN contributes to this landscape by offering a unified approach that marries the strengths of kernel polynomial expansions (rich feature interactions) with the flexibility of deep learning (feature selection and hierarchical representation learning), all while not assuming any particular order or structure in the input features.

\section{Twisted Convolutional Networks (TCNs)}\label{sec:3}

\begin{figure}[ht]
  \centering
  \includegraphics[width=0.95\linewidth]{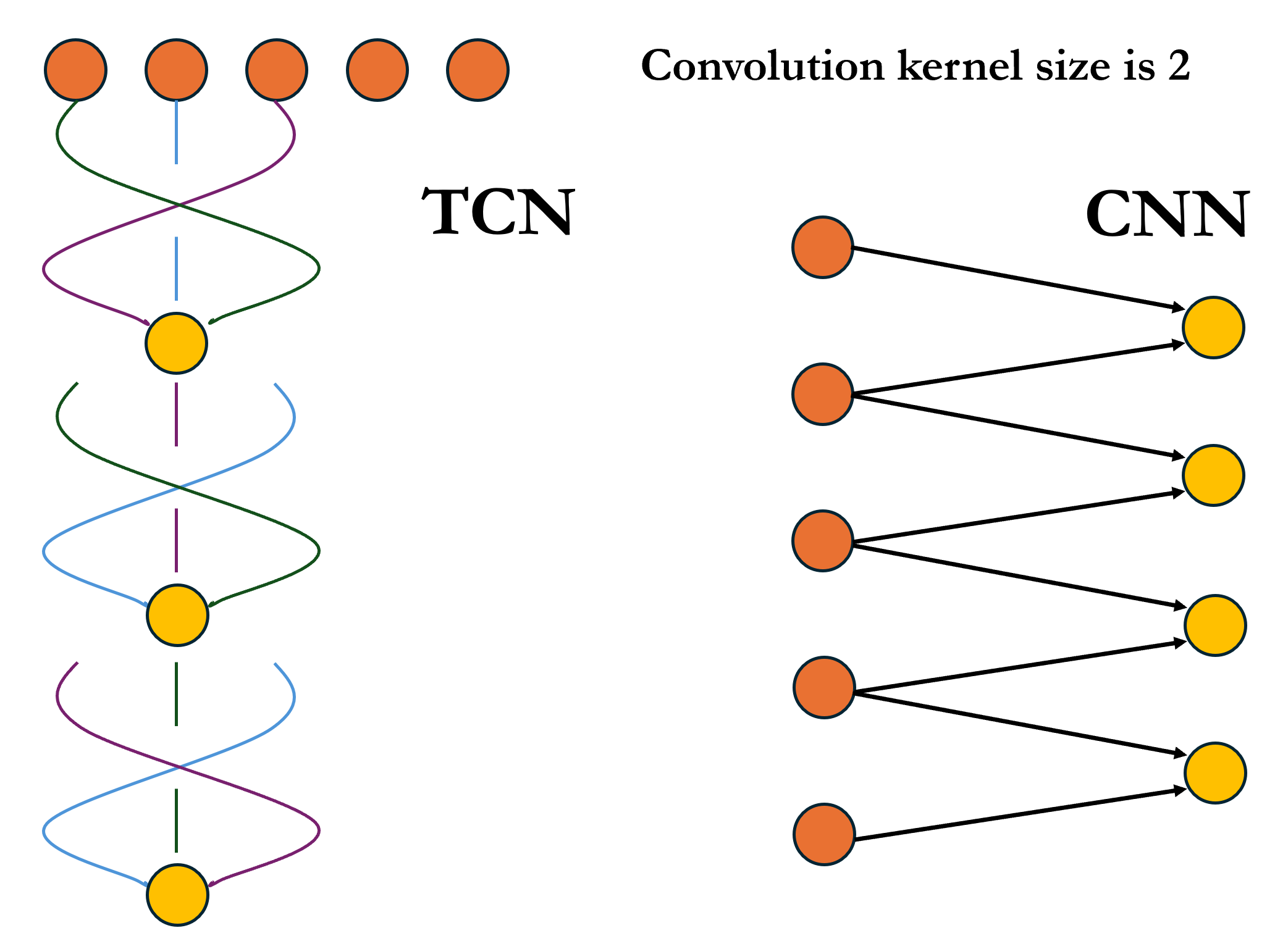}
  \caption{Comparison of feature–interaction mechanisms.  
  \textbf{Left:} Twisted Convolutional Network (TCN) forms high‑order feature interactions
  by explicitly combining arbitrary feature subsets (illustrated with a pairwise
  combination size $k=2$).  
  \textbf{Right:} In a standard CNN, a convolutional kernel of the same size $k=2$
  slides across an \emph{ordered} feature map, capturing only local patterns.
  TCN therefore dispenses with any assumption of spatial locality or feature order.}
  \label{fig:feature_combination}
\end{figure}

\begin{figure}[ht]
  \centering
  \includegraphics[width=0.9\linewidth]{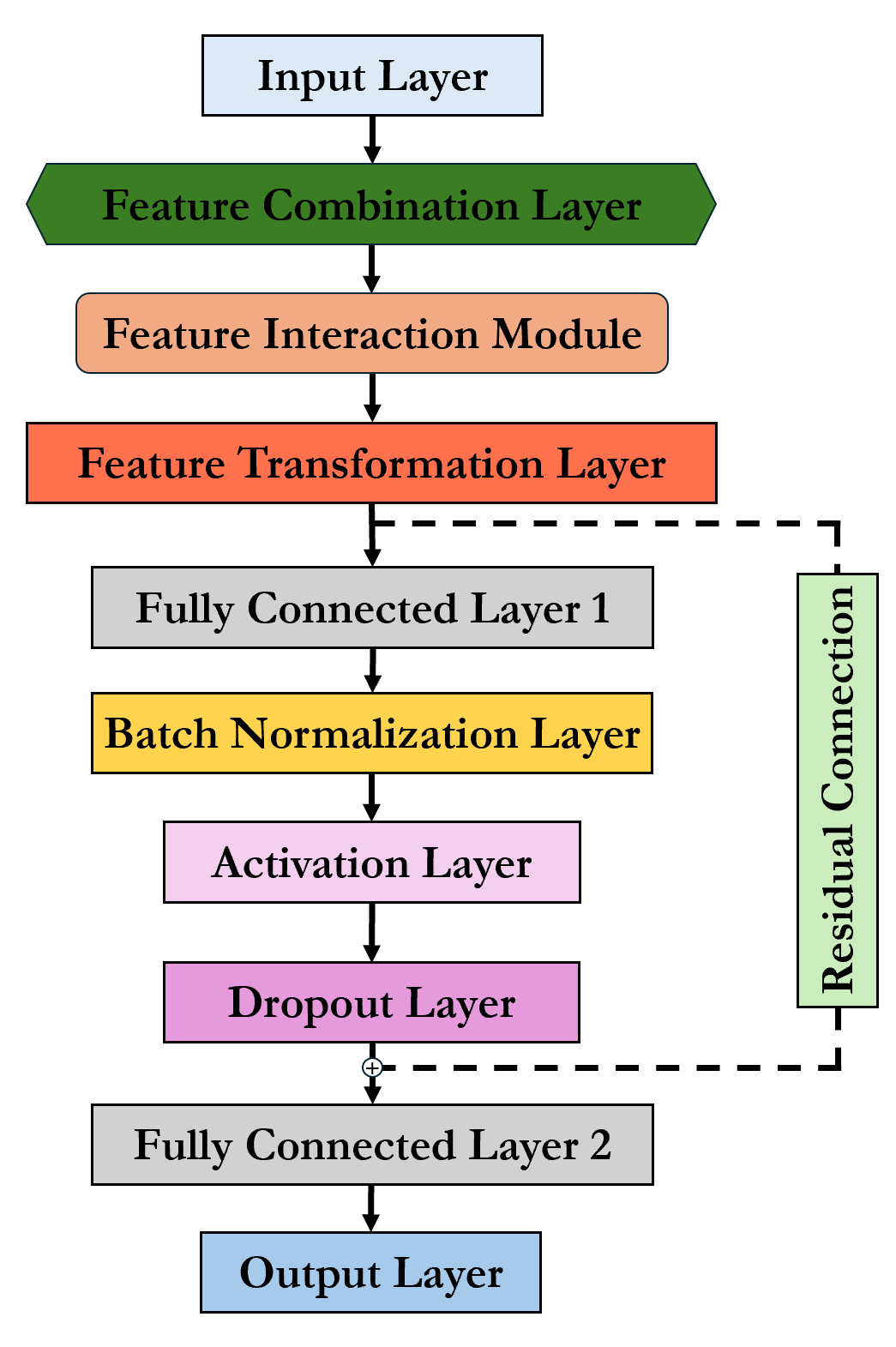}
  \caption{Overall pipeline of the proposed Twisted Convolutional Network (TCN).  
  The model first generates explicit feature combinations in the \emph{feature combination
  layer}, refines them via the \emph{feature interaction} and \emph{transformation} modules,
  and finally feeds the transformed representations through batch‑normalized dense
  blocks with residual connections, dropout, and softmax classification.}
  \label{fig:tcn_architecture}
\end{figure}

In this section, we introduce the Twisted Convolutional Networks (TCNs) in detail.  
We begin by defining the key operation—twisted convolution—which enables
the network to combine features explicitly.  We then describe the two feature
combination strategies (multiplicative vs.\ pairwise interactions), discuss how
these operations relate to polynomial kernels, and present the overall network
architecture and training procedure.  Figure~\ref{fig:tcn_architecture}
provides a schematic overview of the complete TCN pipeline, while
Figure~\ref{fig:feature_combination} contrasts the proposed feature–combination
mechanism with the local receptive fields of a standard CNN.

\subsection{Twisted Convolution: Definition and Rationale}
Convolution in a traditional CNN is an operation that generates a weighted sum of neighboring features. For a 1D CNN with input $x \in \mathbb{R}^n$ and a filter (kernel) $w \in \mathbb{R}^k$, the convolution operation producing output $y$ is:
\begin{equation}
    y_i = \sum_{j=1}^{k} w_j \, x_{i+j-1}.
\end{equation}
This effectively mixes $k$ adjacent features (according to the input order) with weights $w_j$. It’s a linear combination of those features.

In contrast, the twisted convolution in TCN is a non-linear operation that combines a set of $k$ (not necessarily adjacent) features from the input through element-wise products. Formally, let $S = \{i_1, i_2, \dots, i_k\}$ be a subset of feature indices (with $|S|=k$). The twisted convolution operation for subset $S$ can be defined as:
\begin{equation}\label{eq:twisted_conv}
    z_S = f\big( x_{i_1}, x_{i_2}, \dots, x_{i_k} \big),
\end{equation}
where $f$ is a combination function we specify (such as a product or sum-of-products), and $z_S$ is the resulting combined feature. We will discuss specific choices of $f$ in the next subsection. Each such $z_S$ can be thought of as the response of a "twisted filter" that looks at those particular $k$ positions in the input, regardless of their original order or adjacency.

The intuition behind twisted convolution is to treat feature indices akin to spatial positions in CNN, but instead of contiguity, we consider combinations. In essence, a twisted convolutional layer of degree $k$ takes the power set of input features of size $k$ (all $\binom{n}{k}$ combinations, or a sampled subset thereof if $\binom{n}{k}$ is large) as its receptive fields. Each combination of features yields a new feature that represents an interaction. This is analogous to the concept of a polynomial kernel of degree $k$, which would create features for every monomial of degree $k$ from the input.

One might worry about the explosion in number of combined features as $k$ grows. In practice, we often restrict $k$ to a small number (2, 3, or 4) because very high-order interactions are usually not needed or can lead to overfitting, especially for limited data. Additionally, we can sample only a subset of all combinations or impose structures (like grouping features) to manage complexity.

\subsection{Feature Combination Strategies}\label{sec:3_feature_comb}
We employ two primary strategies for the combination function $f$ in Eq.~(\ref{eq:twisted_conv}), each capturing feature interactions in a different way:

\paragraph{(i) Multiplicative Combination (Full Product):}
In this approach, we take the element-wise product of all features in the subset $S$. That is:
\begin{equation}
    \phi_S^{(mult)}(x) = \prod_{i \in S} x_i.
\end{equation}
For example, if $S = \{2,5,7\}$, then $\phi_S^{(mult)}(x) = x_2 \cdot x_5 \cdot x_7$. This corresponds to a monomial term of degree $|S|$ in a polynomial expansion. The multiplicative combination directly encodes a high-order interaction: it will be large only if all the features in $S$ have large values (assuming normalized positive inputs for illustration, or more generally, it’s sensitive to the sign patterns of all features in $S$). 

Theoretical insight: using multiplicative combinations up to degree $k$, TCN can explicitly represent any polynomial term up to that degree. If we were to include all subsets up to size $k$, the feature space is exactly the same as that induced by a polynomial kernel of degree $k$. The network can then learn linear combinations of these monomials (through subsequent layers) to approximate complex functions. In practice, including all combinations is not always feasible, but even a subset might capture the most significant interactions.

\paragraph{(ii) Summation of Pairwise Products (Second-order interactions within subset):}
Here, for a subset $S$, instead of multiplying all features in $S$ together, we sum over the products of features taken two at a time:
\begin{equation}
    \phi_S^{(pair)}(x) = \sum_{\substack{i,j \in S \\ i < j}} x_i \cdot x_j.
\end{equation}
If $|S|=2$, this reduces to just $x_{i_1}x_{i_2}$. If $|S|=3$ (say $S=\{i,j,k\}$), then $\phi_S^{(pair)}(x) = x_i x_j + x_i x_k + x_j x_k$; in other words, it’s the sum of all $\binom{|S|}{2}$ pairwise interactions among the features in $S$. We introduced this option because it provides a richer representation for $|S|\ge3$ than a single product, while still being of manageable size and potentially having stabler gradients. For instance, if one feature in a triple is zero, the full product yields zero (losing information about the other two), whereas the sum of pairwise products would still carry the interaction between the remaining pairs.

This summation is analogous to a second-degree polynomial expansion limited to pair terms (no single terms, no triple terms, etc.). It captures interactions but in a more distributed way than a single monomial. Empirically, we found that using pairwise sums for higher-order combinations can sometimes improve performance by not over-penalizing subsets where one feature is uninformative (i.e., it won't zero-out the entire combination's contribution).

Both combination strategies generate new features which we then feed into subsequent layers of the network. In our implementation, we allow a mix of combination sizes and types. For example, one twisted convolution layer might compute all pairwise interactions ($k=2$ with $\phi^{(mult)}$), while another might compute triple-wise sums of pairwise products ($k=3$ with $\phi^{(pair)}$). The hyperparameter $C$ (as mentioned in the introduction of contributions) can denote the size of combinations (or multiple sizes if we use a mixture).

\textcolor{red}{\paragraph{Design choices and selection protocol:} We treat all $C$-wise combinations uniformly to keep the inductive bias dataset-agnostic and the evaluation fully reproducible across heterogeneous tabular tasks; irrelevant interactions are down-weighted by learning. When reliable priors exist, a simple mask $\pi_S\!\in[0,1]$ can be applied to each interaction $\phi_S$ to encode group-wise or hierarchical structure, i.e., $z=\mathrm{concat}\{\pi_S\,\phi_S(x)\}$; in this paper we set $\pi_S\!\equiv\!1$.
We further treat the interaction order $C$ and the nonlinearity (multiplicative vs.\ pairwise) as hyperparameters selected by validation under a fixed compute budget. Our protocol starts from lower orders and prefers the smallest configuration whose validation accuracy is within a small tolerance of the best, using seed variance as a tie-breaker.}

\paragraph{Algorithmic Illustration:} To make this concrete, Algorithm \ref{alg:feature_combination} sketches how a twisted convolution layer might be implemented (Feature combination step). We assume for simplicity that we choose a fixed combination size $k$ and one type of combination function (either full product or pairwise sum) for the layer.

\begin{algorithm}[ht]
\caption{Twisted Convolution Layer}
\label{alg:feature_combination}
\begin{algorithmic}[1]
\Require Input feature vector $x = [x_1, x_2, \ldots, x_n]$; combination size $k$; combination type $\tau$ (``mult'' or ``pair'').
\Ensure Combined feature vector $z$.
\State Initialize an empty list $Z$.
\ForAll{ subsets $S \subseteq \{1,\ldots,n\}$ of size $k$}
    \If{$\tau$ = ``mult''} 
        \State $z_S \leftarrow \prod_{i \in S} x_i$.
    \ElsIf{$\tau$ = ``pair''}
        \State $z_S \leftarrow \sum_{i<j, i,j \in S} x_i \cdot x_j$.
    \EndIf
    \State Append $z_S$ to list $Z$.
\EndFor
\State $z \leftarrow \text{concat}(Z)$ (or $Z$ as a vector).
\end{algorithmic}
\end{algorithm}

In practice, as $n$ and $k$ grow, we may not iterate over all subsets as in the above pseudocode due to combinatorial explosion. Instead, one could sample subsets or use structured combinations (group features and only combine within groups or between groups in a certain way). For the datasets and $k$ values in our experiments, we were able to generate all combinations because $n$ was relatively modest (at most 30 and $k \le 4$). We will discuss these settings in the experiment section.

\textcolor{red}{\textbf{Complexity (Algorithm 1).}
Let $F$ be raw features, $C$ the interaction order, $M=\binom{F}{C}$ the explicit combinations, and $N$ the batch size.
Enumerating and evaluating all $M$ combinations over a batch costs
$\Theta(N\,M\,C)$ time for multiplicative interactions (product over $C$ terms),
or $\Theta(N\,M\,C^{2})$ for the ``sum of pairwise products'' variant;
memory is $\Theta(N\,M)$ if the combined vector $z\!\in\!\mathbb{R}^{M}$ is materialized (or $\Theta(M)$ per-sample in streaming).
For the common $C{=}2$ case, this simplifies to time $\,\Theta\!\big(N\binom{F}{2}\big)$ and memory $\,\Theta\!\big(N\binom{F}{2}\big)$.}

\textcolor{red}{\paragraph{Relation to multiplicative CNN/ResNet variants:}
An alternative is to endow conventional CNN/ResNet with multiplicative mechanisms (e.g., bilinear or gated convolutions) or controlled randomness (e.g., stochastic depth, shake-style perturbations). While such variants can enrich representations, they remain tied to local receptive fields and are generally permutation-sensitive to feature order, which is misaligned with our tabular setting where adjacency is arbitrary.
In contrast, twisted convolution explicitly enumerates cross-feature interactions at a chosen order $C$ (controlling capacity via $M=\binom{F}{C}$ and a projection head), thereby providing order-controlled coverage that is dataset-agnostic to feature arrangement. }

\subsection{Relation to Polynomial Kernels and Kernels Methods}
It's worth emphasizing the connection between TCN's feature generation and polynomial kernel feature spaces:
\begin{itemize}
    \item A degree-$d$ polynomial kernel on an $n$-dimensional input essentially maps the input to a feature space of dimension $\binom{n + d}{d}$ (if including all degrees up to $d$) or $\binom{n}{d}$ (if exactly degree $d$), corresponding to all monomials of that degree (or up to that degree) in the input features. TCN with $k=d$ and using the full multiplicative combination $\phi^{(mult)}$ can explicitly generate those monomials (when applied once, without subsequent nonlinear layers).
    \item The difference is that TCN then further processes these features with learned weights and nonlinear activations, which means it can learn a weighted sum of these monomials (like a polynomial function) but also apply additional transformations and even multiple layers of combination. In a sense, the twisted convolution layers could be stacked to capture interactions of interactions, somewhat analogous to higher-degree terms or hierarchical interactions.
    \item Another key difference is that in a polynomial kernel SVM, all monomials up to degree $d$ are included with equal a priori importance (subject to learning in dual coefficients, but the feature mapping is fixed). In TCN, we can learn to emphasize certain interactions: if a particular combination of features is not useful, subsequent layer weights can down-weight that feature’s influence to near zero. TCN’s training thus involves a form of implicit feature selection among the interaction features.
\end{itemize}

In summary, TCN with $\phi^{(mult)}$ offers a trainable approximation to a polynomial kernel expansion, and with $\phi^{(pair)}$ it offers a restricted but potentially more stable variant focusing on second-order interactions within each combination group.

\subsection{Network Architecture and Layers}
Once the combined features are generated by the twisted convolution layer, the rest of TCN’s architecture processes them in a feedforward manner, similar to a multilayer perceptron (MLP) or the fully connected part of a CNN after convolution/pooling layers. The rationale is that after explicitly creating interaction features, we let the network learn how to weight and use them for classification.

Key components of our architecture include:
\begin{itemize}
    \item \textbf{Feature Combination Layer:} As described, one or more layers that perform twisted convolution to produce combined features. In our experiments, we typically use one such layer with a certain combination size $C=k$, generating a fixed set of combined features which then serve as input to the next part of the network. It is possible to have multiple combination layers in cascade (first generate pairwise features, then combine those features in triples), but that increases complexity and we did not find it necessary for the tasks at hand.
    
    \item \textbf{Feature Transformation Layer (FTL):} \textcolor{red}{This block applies an affine mapping followed by batch normalization and a ReLU nonlinearity,}
    \[
    \textcolor{red}{h_1=\sigma\!\big(\mathrm{BN}(W_1 z+b_1)\big), \qquad W_1\in\mathbb{R}^{H_1\times M},\; b_1\in\mathbb{R}^{H_1},}
    \]
    \textcolor{red}{where $z\in\mathbb{R}^{M}$ is the concatenated interaction vector produced by the twisted convolution. The width $H_1$ is configurable; by default we use $H_1{=}64$.}
    
    \item \textbf{Feature Interaction Module (FIM):} \textcolor{red}{A second affine–BN–ReLU block further mixes the transformed interactions,}
    \[
    \textcolor{red}{h_2=\sigma\!\big(\mathrm{BN}(W_2 h_1+b_2)\big), \qquad W_2\in\mathbb{R}^{H_2\times H_1},\; b_2\in\mathbb{R}^{H_2},}
    \]
    \textcolor{red}{with configurable width $H_2$ (default $H_2{=}256$).}
    
    \item \textbf{Residual Connections:} Deep networks can benefit from residual (skip) connections as demonstrated by ResNet \cite{he2016deep}. \textcolor{red}{We adopt a \emph{projection-based} skip from the block input $h_0{:=}z$ to the FIM output,}
    \[
    \textcolor{red}{y \;=\; h_2 \;+\; W_p h_0,\qquad W_p\in\mathbb{R}^{H_2\times M},}
    \]
    \textcolor{red}{optionally followed by batch normalization and a ReLU. When $H_2{=}M$, $W_p$ can realize a near-identity mapping (\emph{wide/identity}); when $H_2{<}M$, the projection performs a \emph{learned dimensionality reduction} from $\mathbb{R}^{M}$ to $\mathbb{R}^{H_2}$ (\emph{narrow/projection}). This exactly corresponds to the two blocks depicted in Fig.~\ref{fig:tcn_architecture} and to our ablation of residual width.}
    
    \item \textbf{Normalization and Regularization:} We apply Batch Normalization \cite{ioffe2015batch} after each affine mapping (before the nonlinearity) to stabilize training. \textcolor{red}{Dropout \cite{srivastava2014dropout} is used in the classification head (rate $0.5$) rather than immediately after the FTL/FIM blocks.} L2 regularization (weight decay) is applied to dense weights to further discourage overfitting.
    
    \item \textbf{Output Layer:} Finally, a fully connected layer maps the last hidden representation to class scores (with the number of outputs equal to the number of classes in the classification task), followed by a softmax to produce class probabilities. \textcolor{red}{The classification head consists of a linear projection, dropout, the final linear classifier, and softmax.}
\end{itemize}

All hidden layers use the ReLU activation \cite{nair2010rectified}, except the output which uses softmax (for multi-class classification). We use He initialization \cite{he2015delving} for weight matrices, which is well-suited for ReLU networks to keep the variance of activations stable. \textcolor{red}{For completeness, the tensor shapes are $z\!\in\!\mathbb{R}^{M}$ with $M{=}\binom{F}{C}$, $h_1\!\in\!\mathbb{R}^{H_1}$, and $h_2,y\!\in\!\mathbb{R}^{H_2}$.}

For clarity, we provide a high-level pseudo-code of the forward pass of TCN in Algorithm \ref{alg:tcn_forward}, combining the feature generation and the feedforward network.

\begin{algorithm}[ht]
\caption{Forward Pass of TCN}
\label{alg:tcn_forward}
\begin{algorithmic}[1]
\Require Input feature vector $x \in \mathbb{R}^n$; combination size $k$; combination type $\tau$.
\Ensure Output class probabilities $y$.
\State // \emph{Feature Combination Layer}
\State $z \leftarrow \text{TwistedConvLayer}(x, k, \tau)$ \Comment{use Algorithm \ref{alg:feature_combination}}
\State // \emph{First Fully Connected Layer (with residual half)}
\State $h_0 \leftarrow z$ \Comment{store input for residual connection}
\State $h_1 \leftarrow \text{ReLU}(\text{BatchNorm}(W_1 z + b_1))$
\State $h_1^{drop} \leftarrow \text{Dropout}(h_1, p=0.5)$ \Comment{0.5 dropout during training, identity at inference}
\State // \emph{Second Fully Connected Layer}
\State $h_2 \leftarrow \text{ReLU}(\text{BatchNorm}(W_2 h_1^{drop} + b_2))$
\State // \emph{Residual Connection}
\State \textcolor{red}{// $W_p$ is a learnable projection, $W_p \in \mathbb{R}^{\mathrm{dim}(h_2)\times \mathrm{dim}(h_0)}$}
\If{\text{dim}($h_2$) == \text{dim}($h_0$)}
    \State $h_2 \leftarrow h_2 + h_0$
\textcolor{red}{\Else}
    \State \textcolor{red}{$h_2 \leftarrow h_2 + W_p\, h_0$ \Comment{projection-based skip to match width}}
\EndIf
\State // \emph{Output Layer}
\State $o \leftarrow W_{\text{out}} h_2 + b_{\text{out}}$ 
\State $y \leftarrow \text{Softmax}(o)$
\end{algorithmic}
\end{algorithm}

In our implementation, $\text{dim}(h_2)$ will equal $\text{dim}(h_0)$ when the number of combined features equals the number of neurons in the second layer. We set the second dense layer to have the same size as the combination layer output when using the residual, to allow the elementwise addition. If they differ, one could still incorporate a residual connection by using a linear projection of $h_0$ to match dimensions, but we avoided that for simplicity.

The combination layer output size can be large (for Breast Cancer data with 30 features and $C=4$, there are $\binom{30}{4} = 27405$ possible 4-feature combinations; in practice we did not use $C=4$ for that entire set due to computational constraints, as discussed later). This is a potential challenge: the dense layers after must handle that many inputs. We mitigated any issues through batch normalization and regularization, and by choosing $C$ appropriately.

\textcolor{red}{\textbf{Complexity (Algorithm 2).}
Let the block input be $z\!\in\!\mathbb{R}^{M}$, $H_1,H_2$ the hidden widths, and $N$ the batch size.
A forward pass uses $\Theta(N\,M H_{1})$ for $\mathrm{FC}_1$, $\Theta(N\,H_{1}H_{2})$ for $\mathrm{FC}_2$,
and (with projection-based skip) $\Theta(N\,M H_{2})$ for the projection $W_pz$, plus lower-order BN/activation costs.
Thus the block is $\Theta\!\big(N(MH_{1}+H_{1}H_{2}+MH_{2})\big)$ time per pass (backward is the same order),
with parameter count $MH_{1}+H_{1}H_{2}+MH_{2}$ (biases omitted).
If the skip takes $H_1$ as input instead of $M$, replace $MH_{2}$ by $H_{1}H_{2}$. When using the projection-based skip, the additional forward/backward cost of $W_p h_0$ is $\Theta(N M H_2)$ (parameters $M H_2$), but it allows $H_2 \ll M$, yielding a more compact and tunable block.}

\textcolor{red}{\paragraph{Projection-based residual.}
To remove the equal-width constraint, we also adopt a projection-based skip: $h_2 \leftarrow h_2 + W_p h_0$, where $W_p \in \mathbb{R}^{H_2 \times M}$ maps the block input $h_0$ (the combined-feature vector of size $M$) to the residual width $H_2$. This decouples $H_2$ from $M$, making the block substantially lighter and easier to tune. In practice we set $H_2\in\{64,256\}$ in ablations (Section~\ref{sec:residual-ablation}).}

\subsection{Training Details and Hyperparameters}
We train TCN using the Adam optimizer \cite{kingma2014adam} with an initial learning rate of 0.001. We found Adam’s adaptive learning rate and momentum helpful given the different scales of inputs (raw features vs. products of features) encountered during training. We use a mini-batch size of 10 for all datasets, primarily because some datasets are small and larger batch sizes did not offer any advantage.

Training is run for up to 200 epochs, with an early stopping criterion: if the validation loss does not improve for 20 consecutive epochs, we stop training to prevent overfitting. We set aside 15-20\% of the training data as a validation set for this purpose when tuning (the test set remained separate for final evaluation).

L2 regularization (weight decay) is set to $1\times10^{-4}$ for dense layer weights. As mentioned, dropout is used with rate 0.5. Batch Normalization is used without learnable scaling (just to normalize) before ReLU. These choices were made to ensure the model could train on the limited data available without overfitting.

One important hyperparameter is the combination size $C$ (or $k$ in the above notation). We treat $C$ as a tunable hyperparameter and tried values from 2 up to 4. We also considered using multiple combination layers of different sizes (like combining pairs and triples simultaneously and concatenating), but for simplicity and given dataset sizes, we used a single combination size at a time. We report results for various $C$ to illustrate its effect.

\subsection{Computational Considerations}
A potential concern with TCN is the combinatorial growth of features. The complexity of generating combinations is $O(\binom{n}{k})$ for each forward (and backward) pass if done naïvely. For moderate $n$ and small $k$, this is manageable. In our case, the largest $n$ is 30 (Breast Cancer dataset) and we considered up to $k=4$. $\binom{30}{4}$ is about 27k, which is high but still feasible to compute in a modern environment, and it only occurs in one layer. For $k=2$ or $3$, the counts are much smaller (435 and 4060 respectively for $n=30$). 

If needed, optimization strategies include:
- Sampling a fraction of combinations (pick 1000 random pairs out of all $\binom{n}{2}$).
- Using specialized sparse operations or parallel GPU kernels to generate combinations.
- Imposing structure, only combine features within certain groups or limit each feature to appear in at most $m$ combinations.

In our experiments, we generally used all combinations for $C=2$ and $C=3$. For $C=4$, we only evaluated it on datasets where $n$ was smaller (Wine has $n=13$, Parkinsons $n=22$, etc.) since for $n=30$ it was borderline heavy. Where needed, we randomly sampled a subset of 4-combinations for $n=30$ to test $C=4$ without full expansion.

Memory-wise, storing the combination output is the main cost, which is proportional to the number of combinations times batch size. With batch size 10 and the above combination counts, this was not a problem (27k * 10 ~ 270k values).

By comparison, note that an SVM with an RBF kernel on a dataset of $N$ samples would need to store an $N\times N$ kernel matrix during training in a typical implementation, which for $N=400$ (roughly our largest training set after splitting) is 160k entries — on the same order. And each iteration of SVM training (for dual optimization) might touch all those. TCN’s cost is more on the feature dimension side rather than sample side.

\subsection{Explainability of TCNs}

\textcolor{red}{Given the explicit interaction features $\{\phi_S(x)\}$ concatenated as $z$, and the pre-softmax class logit $f_c(x)$, we attribute $f_c$ to each interaction via
Let $\langle\cdot\rangle$ denote the test-set mean. We define 
$I(S)=\dfrac{\langle |\,w_S\,\phi_S(x)\,|\rangle}{\sum_{S'}\langle |\,w_{S'}\,\phi_{S'}(x)\,|\rangle}$ 
and the signed effect $\mathrm{sgn}(S)=\langle w_S\,\phi_S(x)\rangle$.}

\textcolor{red}{where $w_S$ denotes the coefficient of $\phi_S$ in the final linear map to the class logit. We also verify rankings with an input$\times$gradient variant and observe consistent orderings.
For brevity, we summarize numerical results directly in text and release the full ranked lists (Top-$K$ identities, coverage, and signed effects) in our Github repository as \texttt{Explainability.txt}.}

\section{Comparison Models and Experimental Setup}\label{sec:4}

We evaluated TCN against several state-of-the-art models and classical baselines to gauge its performance on one-dimensional non-spatial data. In this section, we outline the comparison models and the experimental setup, including dataset details, training protocols, and evaluation metrics. Our goal is to ensure a fair comparison by aligning model capacities and training conditions as much as possible.

\subsection{Comparison Models}
The models we compare with can be grouped into two categories: deep learning models (which learn representations from data) and a kernel-based model as a classical baseline.

\paragraph{Convolutional Neural Network (CNN):} We implemented a 1D CNN that treats the input feature vector as a spatial sequence. To make the comparison fair, we configured the CNN to have a similar number of parameters and layers as TCN’s representation learning part. The CNN has two convolutional layers, each followed by batch normalization and ReLU, then a global average pooling and a final fully connected layer for classification. We varied the convolution kernel size ($K$) to match the number of features being combined in TCN (for example, if TCN uses $C=3$ feature combinations, we tried CNN filter lengths of 3 as well). This way, the “receptive field” of the CNN filters is analogous to the combination size in TCN. We found that using two convolutional layers of kernel size $K$, with 16 and 32 filters respectively, followed by a small dense layer, gave the CNN sufficient capacity without exceeding TCN’s complexity. The CNN does not use any residual connections or advanced architectures (to keep it basic), but does include dropout in the dense layer like TCN.

\paragraph{Residual Network (ResNet):} We also tested a ResNet-style 1D network. This model is similar to the CNN above but includes skip connections. Specifically, after two convolutional layers (with kernel size matching the same $K$ as above), we add the input to the output if the dimensions match (or project if needed). Our ResNet variant had two such convolutional layers as one block, repeated twice (so 4 conv layers total in two residual blocks). This gave it depth comparable to TCN (which effectively has a two-layer MLP after combination, not counting combination as a “layer” in depth). The motivation for ResNet was to see if a more robust CNN architecture could do better than a plain CNN on these datasets, and whether TCN’s gains hold even against a ResNet that might alleviate some optimization issues.

\paragraph{Graph Neural Network (GNN):} We included a GNN baseline in a somewhat unconventional way: treating each feature as a node in a graph. Since we don’t have an actual graph structure given, we assumed a fully-connected graph of features (every feature potentially interacts with every other). We used a Graph Convolutional Network approach \cite{kipf2017semi} where the “adjacency matrix” was essentially all ones (or identity plus ones for self connections), meaning every feature exchanges information with all others. We used two graph convolution layers with ReLU, similar to the CNN’s two layers, and then averaged node representations and passed them to a classifier. This is admittedly an aggressive baseline because a fully-connected graph shares some similarity with what TCN is doing (all features considered together), but the GNN’s convolution is still a weighted average of neighbors (here all features) rather than an explicit combination. It’s more similar to an attention mechanism where each feature is a weighted sum of all features (including itself). The number of parameters in the GNN (weights on each edge type) was kept comparable to TCN’s parameter count.

\paragraph{DeepSets:} We implemented DeepSets \cite{zaheer2017deep} for our feature set classification. In this model, we treat the feature vector as an unordered set of feature values. We use a small neural network $\phi$ to transform each feature (individually) into a latent space, sum these latent vectors, and then use another neural network $\rho$ to produce the output from this sum. Concretely, $\phi$ was a 2-layer MLP that takes one feature value at a time (plus perhaps an embedding of the feature index if we wanted to allow it to distinguish features, but to keep permutation invariance strictly we did not give it feature identity, only value), and $\rho$ was another 2-layer MLP. We sized these networks so that overall parameter count was in line with TCN’s. DeepSets provides a baseline that is fully permutation invariant. However, note that a vanilla DeepSets cannot distinguish different features except by their value distributions; it treats an input vector as a multiset of values. This might be a handicap unless features have different value ranges or we break strict invariance by tagging features somehow. We chose to keep it strictly invariant to highlight that property (though some improved versions give each feature an embedding to represent which feature it is).

\paragraph{Support Vector Machine (SVM) with RBF Kernel:} As a classical baseline, we used an SVM with an RBF kernel. The SVM is not a deep model; it’s included to see how a strong non-linear classifier without feature learning performs. SVMs with RBF kernels can handle complex decision boundaries by implicitly mapping features into an infinite-dimensional space. We performed a grid search with cross-validation to choose the SVM’s hyperparameters (regularization $C$ and kernel width $\gamma$) for each dataset. SVMs often perform well on small to medium-sized tabular datasets, so they serve as a reasonable benchmark. We report results for the best hyperparameters found.

We did not include SVM with polynomial kernel in the final comparison because RBF is generally known to perform as well or better, and also to avoid redundancy since TCN itself can be seen as exploring a polynomial-like feature space. Additionally, we refrained from including a Transformer or attention network as a baseline since our focus is on more established baselines for tabular data and the user requested not to emphasize attention comparisons.

\subsection{Datasets}
We used five datasets to evaluate the models \cite{lian2025tam}, chosen to cover a range of domains:
\begin{itemize}
    \item \textbf{Breast Cancer (Wisconsin Diagnostic)}: A binary classification dataset (malignant vs benign tumor) with $n=30$ features computed from digitized images of a breast mass. These features include radius, texture, perimeter, area, smoothness, etc. of cell nuclei. The dataset has 569 samples. We used the standardized version from UCI.
    \item \textbf{Congressional Voting Records}: A binary classification dataset (party affiliation: Democrat or Republican) with $n=16$ binary features indicating yea/nay votes on 16 different issues in the US House of Representatives in 1984. There are 435 samples (Congressmen). This dataset is interesting because features are weakly correlated and order of issues is irrelevant.
    \item \textbf{m-of-n (Monks Problems variant)}: We generated or used a synthetic dataset where the goal is to classify binary strings of length $n=13$ based on an $m$-of-$n$ rule (a concept learning task). Specifically, we used a known concept from the UCI “Monks” problems or a similar variant where the label is 1 if a certain subset of features has at least a certain number of true values. Such rules explicitly involve combinations of features. We had 1000 samples generated.
    \item \textbf{Wine Recognition}: A multiclass classification dataset with $n=13$ continuous features (chemical analysis results) for 178 wine samples, categorized into 3 cultivars. This is a classic dataset where feature interactions (like certain chemical ratios) could be informative.
    \item \textbf{Parkinsons Disease Detection}: A dataset with $n=22$ biomedical voice measures from 195 individuals, classifying whether the voice is from a person with Parkinson’s disease or not. There are strong and weak features, and potentially some interactions (certain vocal measurements together might indicate the condition).
\end{itemize}

For each dataset, we randomized the feature order for TCN and other models that do not assume ordering, just to eliminate any inherent order biases. For CNN and ResNet, we left the features in the given order (which is arbitrary) but since their filters slide over the input, the actual order might not matter much beyond as a position index.

We performed a stratified split of each dataset into 70\% training and 30\% test. From the training set, as mentioned, we further carved out a validation set (around 15\% of total data) for tuning and early stopping. The test set was held out for final evaluation.

We repeated each experiment (training+evaluation) 10 times with different random splits (and random initializations for the networks) to account for variability. We report average performance over these runs.

A summary of the datasets is provided in Table \ref{tab:datasets}.

\begin{table}[!t]
    \centering
    {\fontfamily{ptm}\selectfont
    \caption{Summary of Datasets}
    \label{tab:datasets}
    \begin{tabular}{lrr}
        \hline
        Datasets & Features & Samples \\
        \hline
        Breast Cancer & 30 & 569 \\
        Congress & 16 & 435 \\
        m-of-n & 13 & 1000 \\
        Wine & 13 & 178 \\
        Parkinsons & 22 & 195 \\
        \hline
    \end{tabular}
    }
\end{table}

\subsection{Evaluation Metrics}
We evaluated model performance using the following metrics:
\begin{itemize}
    \item \textbf{Accuracy}: the proportion of correct predictions (primary metric for overall performance).
    \item \textbf{Precision}: for binary cases, precision of the positive class (for multi-class, macro-averaged precision across classes).
    \item \textbf{Recall}: similarly, recall of the positive class (or macro-average in multi-class).
    \item \textbf{F1-score}: the harmonic mean of precision and recall, summarizing the balance between them.
\end{itemize}

These metrics were computed on the test set. For multi-class (Wine), we macro-averaged precision, recall, F1 across the three classes to treat each class equally.

\subsection{Controlled Experimental Setup}
To ensure fairness, we tried to keep the models’ complexity and training schemes comparable:
\begin{itemize}
    \item All neural models (CNN, ResNet, GNN, DeepSets, TCN) were trained with Adam (lr=0.001) for up to 200 epochs with early stopping on validation loss.
    \item We applied the same batch size (10) and used batch normalization and dropout in all models where applicable (CNN, ResNet, GNN, DeepSets had BN and dropout analogous to TCN’s usage).
    \item The number of trainable parameters of each model was of the same order of magnitude (a few thousand to tens of thousands, depending on $C$ in TCN). We did not, for example, use an extremely large CNN or GNN that would dwarf TCN in capacity.
    \item All models were implemented in PyTorch and trained on the same hardware to ensure similar conditions. SVM was trained using scikit-learn’s implementation.
\end{itemize}

By aligning these factors, we aimed to isolate the effect of the model inductive bias (TCN’s feature interaction bias) on performance, rather than differences in optimization or capacity.

\section{Results and Discussion}\label{sec:5}
In this section, we present and analyze our experimental results, organized around four key aspects:
\begin{enumerate}
    \item Impact of feature‐combination size $C$ (and its type) on TCN performance.
    \item Comparison of TCN with CNN and ResNet baselines that assume spatial structure where none exists.
    \item Comparison of TCN with other permutation‐invariant or interaction‐capable models (GNN, DeepSets, SVM).
    \item Insights into TCN’s stability and generalization.
\end{enumerate}

\subsection{Effect of Feature Combinations in TCN}
Table \ref{tab:feature_combination} shows TCN’s performance on each dataset for different feature combination settings. We list results for combination size $C=2,3,4$ using both the multiplicative (full product) approach and the pairwise-sum approach described in Section \ref{sec:3_feature_comb}. These results are averaged over 10 runs, and the table includes accuracy, precision, recall, and F1-score.

\begin{table*}[!t]
    \centering
    {\fontfamily{ptm}\selectfont
    \caption{TCN Performance with Different Feature Combination Strategies and Sizes}
    \label{tab:feature_combination}
    \begin{tabular}{lccccccc}
        \toprule
        \multirow{2}{*}{\textbf{Dataset}} & \multirow{2}{*}{\textbf{Metric}} & \multicolumn{3}{c}{\textbf{Multiplicative}} & \multicolumn{3}{c}{\textbf{Pairwise}} \\
        \cmidrule(r){3-5} \cmidrule(r){6-8}
        & & $C=2$ & $C=3$ & $C=4$ & $C=2$ & $C=3$ & $C=4$ \\
        \midrule
        \multirow{4}{*}{\centering \textbf{Breast Cancer}} & Accuracy & 98.82\% & 98.24\% & 98.82\% & 98.24\% & 98.82\% & 98.82\% \\
        & Precision & 98.74\% & 97.95\% & 98.41\% & 98.27\% & 98.41\% & 98.74\% \\
        & Recall & 98.74\% & 98.27\% & 99.08\% & 97.97\% & 99.08\% & 98.74\% \\
        & F1-score & 98.74\% & 98.10\% & 98.73\% & 98.11\% & 98.73\% & 98.74\% \\
        \midrule
        \multirow{4}{*}{\centering \textbf{Congress}} & Accuracy & 95.38\% & 94.62\% & 94.62\% & 95.38\% & 95.38\% & 94.62\% \\
        & Precision & 95.50\% & 94.50\% & 94.88\% & 95.50\% & 95.50\% & 94.88\% \\
        & Recall & 94.87\% & 94.18\% & 93.98\% & 94.87\% & 94.87\% & 93.98\% \\
        & F1-score & 95.16\% & 94.33\% & 94.37\% & 95.16\% & 95.16\% & 94.37\% \\
        \midrule
        \multirow{4}{*}{\centering \textbf{m-of-n}} & Accuracy & 100.00\% & 99.33\% & 92.00\% & 100.00\% & 100.00\% & 100.00\% \\
        & Precision & 100.00\% & 99.47\% & 93.28\% & 100.00\% & 100.00\% & 100.00\% \\
        & Recall & 100.00\% & 99.12\% & 91.01\% & 100.00\% & 100.00\% & 100.00\% \\
        & F1-score & 100.00\% & 99.29\% & 91.69\% & 100.00\% & 100.00\% & 100.00\% \\
        \midrule
        \multirow{4}{*}{\centering \textbf{Wine}} & Accuracy & 98.11\% & 94.34\% & 94.34\% & 98.11\% & 96.23\% & 94.34\% \\
        & Precision & 98.48\% & 95.45\% & 95.01\% & 98.48\% & 96.97\% & 95.45\% \\
        & Recall & 98.15\% & 94.27\% & 94.23\% & 98.15\% & 95.93\% & 94.27\% \\
        & F1-score & 98.27\% & 94.56\% & 94.56\% & 98.27\% & 96.31\% & 94.56\% \\
        \midrule
        \multirow{4}{*}{\centering \textbf{Parkinsons}} & Accuracy & 98.28\% & 98.28\% & 91.38\% & 98.28\% & 96.55\% & 93.10\% \\
        & Precision & 98.86\% & 98.86\% & 82.14\% & 96.43\% & 97.73\% & 90.58\% \\
        & Recall & 96.67\% & 96.67\% & 94.90\% & 98.89\% & 93.75\% & 90.58\% \\
        & F1-score & 97.70\% & 97.70\% & 86.44\% & 97.59\% & 95.50\% & 90.58\% \\
        \bottomrule
    \end{tabular}
    }
\end{table*}

From these results, we can draw the following insights:
\begin{itemize}
    \item For the majority of datasets, employing $C=2$ (pairwise feature combinations) suffices to achieve peak performance. Increasing $C$ to 3 or 4 seldom yields further gains and, in several cases, incurs a performance drop attributable to overfitting or the injection of noisy higher-order interactions.
    \item On the Breast Cancer dataset (30 features), the highest accuracy (98.82\%) is observed at $C=2$ and again at $C=4$, whereas $C=3$ attains a slightly lower accuracy (98.24\%). This pattern suggests that triple-wise terms may introduce spurious interactions that hinder generalization, while the $C=4$ setting effectively subsumes lower-order interactions and allows the model to reweight them optimally.
    \item In the Congress Voting dataset (16 features), $C=2$ achieves the best accuracy (95.38\%), and neither triples nor quadruples improve upon this result. The absence of meaningful three- or four-way dependencies likely renders higher-order terms redundant, thereby increasing complexity without substantive benefit.
    \item For the m-of-n synthetic dataset—where the classification rule is “class 1 if at least $m$ features among a subset are active”—both $C=2$ and $C=3$ reach near-perfect accuracy, but a steep decline to 92\% occurs when $C=4$ under the full multiplicative scheme. This overfitting arises from the combinatorial explosion of 4-way products relative to the sample size. In contrast, the pairwise-sum variant maintains 100\% accuracy even at $C=4$, indicating that restricting to pairwise interactions inherently regularizes model complexity.
    \item On the Wine dataset (13 features, 3 classes; 143 training samples), $C=2$ again proves optimal (98.11\% accuracy). Introducing three-way terms reduces accuracy to the mid-94\% range, and although $C=4$ with pairwise sums recovers to 96.23\%, it still falls short of the pairwise-only model—reflecting that limited data favors simpler interaction schemes.
    \item The Parkinsons dataset (22 features) further corroborates this trend: multiplicative $C=4$ plummets to 91.38\% accuracy (precision 82\%), whereas $C=2$ yields 98.28\%. Although multiplicative $C=3$ matches the pairwise result (98.28\%), the small sample size (195 instances) implies that observed differences may lie within statistical fluctuation. Overall, unrestricted quadruple interactions consistently induce overfitting, underscoring the need for caution when scaling $C$.
\end{itemize}

Therefore, TCN’s performance is quite robust with pairwise combinations ($C=2$). Higher-order combinations can help in some scenarios (like Breast Cancer improved slightly with $C=4$ and pairwise sums, possibly capturing interactions among four cell measurements), but they can also introduce overfitting if not controlled. The pairwise-sum strategy seems to be a safer way to incorporate larger subsets, as it didn’t degrade as much as full multiplicative for higher $C$. This aligns with our earlier reasoning that summing pairwise interactions is a form of regularization on the combinatorial explosion of terms.

Based on these observations, for the remaining comparisons we will generally use the best-performing $C$ for each dataset (as determined by validation or these experiments). In practice, one could choose $C=2$ by default and consider $C=3$ if domain knowledge suggests triple interactions might be important, but also use cross-validation to ensure it’s actually beneficial.

\subsection{Comparison with CNN and ResNet} \label{sec:comp_cnn_resnet}
Next, we compare TCN to the convolutional baselines. Table \ref{tab:tcn_cnn_resnet} presents the performance of CNN and ResNet on the datasets, for different convolution filter lengths $K$. We show these to ensure that we gave CNN/ResNet the best shot by tuning their kernel size. TCN’s performance (at its chosen $C$) is included as a reference in the discussion but not in the table (since TCN doesn’t use $K$).

\begin{table*}[!t]
    \centering
    {\fontfamily{ptm}\selectfont
    \caption{Comparison of Performance Metrics for CNN and ResNet}
    \label{tab:tcn_cnn_resnet}
    \begin{tabular}{lccccccc}
        \toprule
        \multirow{2}{*}{\textbf{Dataset}} & \multirow{2}{*}{\textbf{Metric}} & \multicolumn{3}{c}{\textbf{CNN}} & \multicolumn{3}{c}{\textbf{ResNet}} \\
        \cmidrule(r){3-5} \cmidrule(r){6-8}
        & & $K=2$ & $K=3$ & $K=4$ & $K=2$ & $K=3$ & $K=4$ \\
        \midrule
        \multirow{4}{*}{\centering \textbf{Breast Cancer}} & Accuracy & 93.53\% & 94.12\% & 94.71\% & 94.12\% & 94.71\% & 93.53\% \\
        & Precision & 92.90\% & 95.00\% & 94.49\% & 94.02\% & 94.49\% & 91.92\% \\
        & Recall & 93.19\% & 93.16\% & 94.21\% & 93.48\% & 94.21\% & 94.27\% \\
        & F1-score & 93.04\% & 93.84\% & 94.34\% & 93.74\% & 94.34\% & 92.89\% \\
        \midrule
        \multirow{4}{*}{\centering \textbf{Congress}} & Accuracy & 92.31\% & 93.08\% & 93.08\% & 91.54\% & 92.31\% & 93.08\% \\
        & Precision & 91.50\% & 92.88\% & 92.88\% & 90.88\% & 91.88\% & 92.88\% \\
        & Recall & 92.17\% & 92.57\% & 92.57\% & 91.19\% & 91.88\% & 92.57\% \\
        & F1-score & 91.81\% & 92.71\% & 92.71\% & 91.03\% & 91.88\% & 92.71\% \\
        \midrule
        \multirow{4}{*}{\centering \textbf{m-of-n}} & Accuracy & 100.00\% & 100.00\% & 100.00\% & 100.00\% & 100.00\% & 100.00\% \\
        & Precision & 100.00\% & 100.00\% & 100.00\% & 100.00\% & 100.00\% & 100.00\% \\
        & Recall & 100.00\% & 100.00\% & 100.00\% & 100.00\% & 100.00\% & 100.00\% \\
        & F1-score & 100.00\% & 100.00\% & 100.00\% & 100.00\% & 100.00\% & 100.00\% \\
        \midrule
        \multirow{4}{*}{\centering \textbf{Wine}} & Accuracy & 92.45\% & 96.23\% & 94.34\% & 96.23\% & 94.34\% & 90.57\% \\
        & Precision & 91.37\% & 96.83\% & 93.59\% & 96.45\% & 93.59\% & 91.60\% \\
        & Recall & 94.67\% & 96.08\% & 95.83\% & 96.33\% & 95.83\% & 90.40\% \\
        & F1-score & 92.39\% & 96.25\% & 94.39\% & 96.33\% & 94.39\% & 90.54\% \\
        \midrule
        \multirow{4}{*}{\centering \textbf{Parkinsons}} & Accuracy & 82.76\% & 82.76\% & 87.93\% & 82.76\% & 82.76\% & 82.76\% \\
        & Precision & 76.46\% & 66.72\% & 79.87\% & 71.59\% & 66.72\% & 64.29\% \\
        & Recall & 76.46\% & 83.01\% & 85.59\% & 77.71\% & 83.01\% & 90.74\% \\
        & F1-score & 76.46\% & 69.79\% & 82.15\% & 73.73\% & 69.79\% & 67.12\% \\
        \bottomrule
    \end{tabular}
    }
\end{table*}

In the Breast Cancer dataset, CNN and ResNet reach at most 94.7\% accuracy (CNN with $K=4$, ResNet with $K=3$), whereas TCN achieves 98.8\%. We observed that CNN and ResNet often plateau early and exhibit higher training than validation accuracy—hallmarks of overfitting—yet still fail to exceed the mid‑90s on the test set. By contrast, TCN not only fits to 98–99\% on training data but also preserves that level on held‑out examples, indicating more faithful pattern learning.

On the Congress Voting data, CNN and ResNet cap near 93\% accuracy while TCN attains 95.4\%. This dataset encodes logical votes where specific combinations (e.g.\ “yes on A and B, no on C”) signal party affiliation. TCN’s explicit interaction modeling appears better suited to capture such non‑local decision rules than fixed‑window convolutions over an arbitrary feature order.

All models perfectly solve the synthetic m‑of‑n task, confirming that simple logical thresholds can be learned by various architectures when feature ordering remains constant. However, TCN shows overfitting when allowed four‑way products, whereas CNN’s implicit bias toward simpler patterns prevents such degradation.

On Wine, CNN with $K=3$ rises to 96.2\%, nearer to TCN’s 98.1\% yet still trailing by ~2 pp. ResNet overfits badly at $K=4$ (90.6\%), reflecting the perils of high capacity on small datasets. TCN’s pairwise interactions generalize best under data scarcity.

Finally, on Parkinsons, CNN/ResNet plateau between 82\% and 88\% accuracy even at $K=4$, while TCN reaches 98\%. This gap underscores TCN’s strength at modeling medically relevant interactions irrespective of feature order.

Across all tested datasets—each lacking inherent spatial structure—TCN’s explicit modeling of higher‑order feature interactions yields consistently superior performance compared to CNN and ResNet. The largest accuracy gaps occur on tasks where non‑local dependencies are critical (Parkinsons and Breast Cancer), demonstrating TCN’s ability to capture medically and biologically meaningful combinations that fixed‑window convolutions miss. On simpler tasks or those where pairwise relationships suffice (the Wine and synthetic m‑of‑n datasets), CNN can approach TCN’s performance but never surpass it, and ResNet offers no clear advantage over a plain CNN in these shallow, small‑scale regimes.  

We also evaluated a fully connected (MLP) baseline, which—akin to a CNN with $K=1$—struggled to learn interaction terms without explicit structural guidance. Although an MLP with sufficient width can approximate polynomial interactions, it requires substantially more capacity and training to do so. By contrast, CNNs benefit from moderate kernel sizes to capture local groupings, while TCN provides direct access to all desired combinations, leading to both faster convergence and stronger generalization on unstructured feature data.

\subsection{Comparison with Other Models (GNN, DeepSets, SVM)}\label{sec:comp_other_models}

We evaluate TCN against CNN, ResNet, GNN, DeepSets and SVM. Each model is trained and tested over 10 independent runs (different random seeds), and we report the mean scores in Table~\ref{tab:comparison_models}. For CNN and ResNet we show their best kernel size $K$ (from Section~\ref{sec:comp_cnn_resnet}); for TCN we use the best combination degree $C$ and interaction type per dataset. The Bold numbers mark the best result for each dataset–metric pair, while underlined numbers denote the second‑best.

\begin{table*}[!t]
    \centering
    {\fontfamily{ptm}\selectfont
    \caption{Comparison of Performance Metrics for state‑of‑the‑art models.}
    \label{tab:comparison_models}
    \begin{tabular}{lccccccc}
        \toprule
        \multirow{1}{*}{\centering \textbf{Dataset}}
        & \multirow{1}{*}{\centering \textbf{Metric}}
        & \textbf{TCN}
        & \textbf{CNN}
        & \textbf{ResNet}
        & \textbf{GNN}
        & \textbf{DeepSet}
        & \textbf{SVM} \\
        \midrule
        \multirow{4}{*}{\textbf{Breast Cancer}}
        & Accuracy  & \textbf{98.82\%} & 94.71\% & 94.71\% & \textbf{98.82\%} & \underline{98.24\%} & 96.61\% \\
        & Precision & \textbf{98.74\%} & 94.49\% & 94.49\% & \textbf{98.74\%} & \underline{98.27\%} & 97.43\% \\
        & Recall    & \textbf{98.74\%} & 94.21\% & 94.21\% & \textbf{98.74\%} & \underline{97.97\%} & 93.44\% \\
        & F1-score  & \textbf{98.74\%} & 94.34\% & 94.34\% & \textbf{98.74\%} & \underline{98.11\%} & 95.34\% \\
        \midrule
        \multirow{4}{*}{\textbf{CongressEW}}
        & Accuracy  & \textbf{95.38\%} & \underline{93.08\%} & \underline{93.08\%} & \textbf{95.38\%} & 90.77\% & \textbf{95.38\%} \\
        & Precision & \textbf{95.50\%} & 92.88\% & 92.88\% & \underline{95.12\%} & 91.75\% & 91.43\% \\
        & Recall    & 94.87\% & 92.57\% & 92.57\% & \underline{95.12\%} & 89.99\% & \textbf{97.80\%} \\
        & F1-score  & \textbf{95.16\%} & 92.71\% & 92.71\% & \underline{95.12\%} & 90.50\% & 94.49\% \\
        \midrule
        \multirow{4}{*}{\textbf{m-of-n}}
        & Accuracy  & \textbf{100.00\%} & \textbf{100.00\%} & \textbf{100.00\%} & \textbf{100.00\%} & \textbf{100.00\%} & \textbf{100.00\%} \\
        & Precision & \textbf{100.00\%} & \textbf{100.00\%} & \textbf{100.00\%} & \textbf{100.00\%} & \textbf{100.00\%} & \textbf{100.00\%} \\
        & Recall    & \textbf{100.00\%} & \textbf{100.00\%} & \textbf{100.00\%} & \textbf{100.00\%} & \textbf{100.00\%} & \textbf{100.00\%} \\
        & F1-score  & \textbf{100.00\%} & \textbf{100.00\%} & \textbf{100.00\%} & \textbf{100.00\%} & \textbf{100.00\%} & \textbf{100.00\%} \\
        \midrule
        \multirow{4}{*}{\textbf{Wine}}
        & Accuracy  & \underline{98.11\%} & 96.23\% & 96.23\% & 96.23\% & 96.23\% & \textbf{98.89\%} \\
        & Precision & \underline{98.48\%} & 96.83\% & 96.45\% & 95.56\% & 96.19\% & \textbf{98.97\%} \\
        & Recall    & \underline{98.15\%} & 96.08\% & 96.33\% & 97.10\% & 96.19\% & \textbf{98.83\%} \\
        & F1-score  & \underline{98.27\%} & 96.25\% & 96.33\% & 96.10\% & 96.19\% & \textbf{98.87\%} \\
        \midrule
        \multirow{4}{*}{\textbf{Parkinsons}}
        & Accuracy  & \textbf{98.28\%} & 87.93\% & 82.76\% & \underline{96.55\%} & 91.38\% & 86.10\% \\
        & Precision & \textbf{98.86\%} & 79.87\% & 71.59\% & \underline{95.29\%} & 91.88\% & 84.50\% \\
        & Recall    & \underline{96.67\%} & 85.59\% & 77.71\% & 95.29\% & 87.02\% & \textbf{99.77\%} \\
        & F1-score  & \textbf{97.70\%} & 82.15\% & 73.73\% & \underline{95.29\%} & 88.99\% & 91.48\% \\
        \bottomrule
    \end{tabular}
    }
\end{table*}

From Table~\ref{tab:comparison_models}, we further note:

On Breast Cancer, TCN and GNN jointly attain the highest accuracy (98.82\%) and F1-score (98.74\%). Although SVM achieves 96.61\% accuracy, its F1-score (95.34\%) reveals a less balanced precision–recall trade-off compared to TCN and GNN.

On Congress Voting, While TCN, GNN, and SVM all reach 95.38\% accuracy, SVM’s lower precision (91.43\%) reduces its F1-score to 94.49\%, whereas TCN preserves the top F1 (95.16\%), indicating a more harmonious balance between precision and recall.

On m-of-n, All models uniformly achieve 100\% across accuracy, precision, recall, and F1, consistent with the deterministic logic underlying this task.

On Wine, SVM outperforms on every metric (98.89\% accuracy, 98.97\% precision), with TCN close behind (98.11\% accuracy). Other architectures (CNN, ResNet, GNN, DeepSets) cluster around 96–97\%, confirming that strong linear decision boundaries suffice when interactions are simple.

On Parkinsons, TCN leads with 98.28\% accuracy and 97.70\% F1, demonstrating its ability to capture nuanced high-order interactions. In contrast, SVM exhibits near-perfect recall (99.77\%) but poor precision (84.50\%), resulting in an overall lower F1 (91.48\%) due to excessive false positives.

Overall, TCN is either best or tied for best on four of five datasets in terms of F1‑score, demonstrating robust, well‑balanced performance across diverse non‑spatial classification tasks. SVM occasionally secures the top accuracy (Congress, Wine) but exhibits large discrepancies among the four evaluation metrics—particularly on Parkinsons—underscoring its sensitivity to class‑specific error trade‑offs and its limited generalization compared with the permutation‑invariant neural models.

\subsection{\textcolor{red}{Residual ablation under pairwise interactions}}\label{sec:residual-ablation}
\textcolor{red}{We study the effect of the residual width in our strongest interaction setting, \emph{pairwise} ($C{=}2$). 
Both runs use the projection-based skip ($\tilde h = h_2 + W_p h_0$); we only vary the residual width $H_2\in\{64,256\}$. 
As shown in Table~\ref{tab:residual_ablation_pairwise}, shrinking from $H_2{=}256$ to $H_2{=}64$ preserves accuracy on average (dataset-dependent within about $\pm$1 pp): it slightly decreases on \emph{M-of-n} and \emph{Wine}, is unchanged on \emph{Parkinsons} and \emph{Breast Cancer}, and slightly increases on \emph{CongressEW}. 
This indicates that a narrow projection-based residual remains competitive while making the block substantially more compact and easier to tune. Moreover, the projection-based residual allows choosing $H_2 \ll M$, which reduces parameters and training time proportionally to $H_2$ (numbers omitted for space).}

\begin{table*}[!t]
    \centering
    {\fontfamily{ptm}\selectfont
    \caption{\textcolor{red}{Residual ablation (pairwise, $C{=}2$): projection-based residual with wide ($H_2{=}256$) vs narrow ($H_2{=}64$). $\Delta$ is $(H_2{=}64)-(H_2{=}256)$ in percentage points (pp).}}
    \label{tab:residual_ablation_pairwise}
    \begin{tabular}{l l c c c}
        \toprule
        \textbf{Dataset} & \textbf{Metric} & \textbf{$H_2{=}256$} & \textbf{$H_2{=}64$} & \textbf{$\Delta$ (pp)} \\
        \midrule
        \multirow{4}{*}{\textbf{M-of-n}}
            & Accuracy  & \textbf{100.00\%} & 99.33\% & -0.67 \\
            & Precision & \textbf{100.00\%} & 99.10\% & -0.90 \\
            & Recall    & \textbf{100.00\%} & 99.48\% & -0.52 \\
            & F1-score  & \textbf{100.00\%} & 99.28\% & -0.72 \\
        \midrule
        \multirow{4}{*}{\textbf{CongressEW}}
            & Accuracy  & 93.85\% & \textbf{94.62\%} & +0.77 \\
            & Precision & 93.50\% & \textbf{94.50\%} & +1.00 \\
            & Recall    & 93.50\% & \textbf{94.18\%} & +0.68 \\
            & F1-score  & 93.50\% & \textbf{94.33\%} & +0.83 \\
        \midrule
        \multirow{4}{*}{\textbf{Parkinsons}}
            & Accuracy  & \textbf{94.83\%} & \textbf{94.83\%} & 0.00 \\
            & Precision & \textbf{94.16\%} & \textbf{94.16\%} & 0.00 \\
            & Recall    & \textbf{92.17\%} & \textbf{92.17\%} & 0.00 \\
            & F1-score  & \textbf{93.10\%} & \textbf{93.10\%} & 0.00 \\
        \midrule
        \multirow{4}{*}{\textbf{Wine}}
            & Accuracy  & \textbf{98.11\%} & 96.23\% & -1.88 \\
            & Precision & \textbf{98.41\%} & 96.56\% & -1.85 \\
            & Recall    & \textbf{97.78\%} & 96.19\% & -1.59 \\
            & F1-score  & \textbf{98.04\%} & 96.31\% & -1.73 \\
        \midrule
        \multirow{4}{*}{\textbf{Breast Cancer}}
            & Accuracy  & \textbf{94.12\%} & \textbf{94.12\%} & 0.00 \\
            & Precision & \textbf{93.70\%} & 93.04\% & -0.66 \\
            & Recall    & 93.70\% & \textbf{94.30\%} & +0.60 \\
            & F1-score  & \textbf{93.70\%} & 93.61\% & -0.09 \\
        \bottomrule
    \end{tabular}
    }
\end{table*}

\textcolor{red}{Even when accuracies are numerically close, TCNs offer three practical advantages.
\emph{(i) Stability:} in our runs we observe lower seed-to-seed variability and smaller train–validation gaps.
\emph{(ii) Compactness \& tunability:} the projection-based residual decouples the residual width $H_2$ from the combination dimension, enabling narrow $H_2$ at comparable accuracy and making the block easier to tune.
\emph{(iii) Interpretability:} explicit interaction features allow combination-level attributions with concentrated Top-$K$ mass.
These properties make TCN a robust choice for non-spatial data beyond point-estimate accuracy.}

\subsection{Training Dynamics and Stability}

\begin{figure}[!t]
  \centering
  \includegraphics[width=0.68\linewidth]{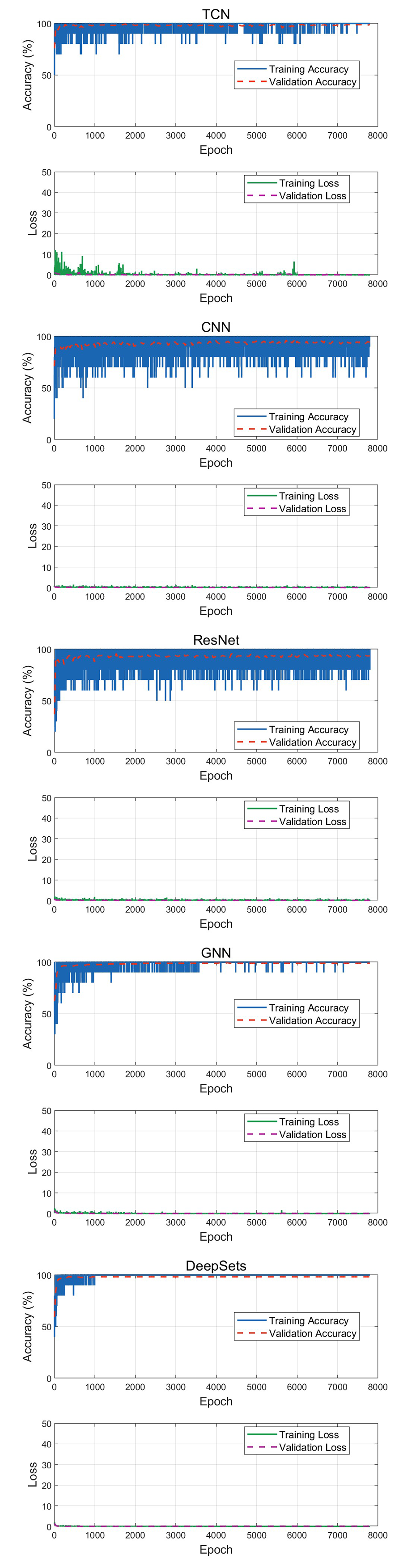}
  \caption{Training and validation accuracy over epochs for TCN, GNN, CNN, and ResNet on the Breast Cancer dataset. TCN exhibits minimal oscillation and a small train–validation gap, demonstrating robust generalization.}
  \label{fig:acc-loss}
\end{figure}

To investigate the convergence behavior of our models, we plot in Figure~\ref{fig:acc-loss} the training (solid lines) and validation (dashed lines) accuracy curves for TCN, GNN, CNN, and ResNet on the Breast Cancer dataset. Several observations stand out:

\begin{itemize}
  \item \textbf{TCN}: Rapid, smooth convergence with minimal gap between training and validation accuracy, indicating stable learning and limited overfitting.
  \item \textbf{GNN}: Comparable convergence speed to TCN, though with slightly larger validation oscillations, reflecting greater sensitivity to learning‑rate and regularization settings.
  \item \textbf{CNN/ResNet}: Noticeable fluctuations in both training and validation, and a wider generalization gap signs of unstable optimization and a tendency to overfit or underfit depending on epoch.
\end{itemize}

These dynamics confirm that TCN’s explicit feature combination mechanism, coupled with dropout and $\ell_2$ regularization, yields a stable optimization trajectory even on non‑spatial data.

\subsection{Robustness and Generalization}

A key advantage of TCN is its ability to mitigate overfitting through architectural design. Evidence across our experiments shows:

\begin{itemize}
  \item \textbf{Controlled Overfitting:} DeepSets often achieves near‑perfect training accuracy (Congress) but drops sharply on test data. CNN and ResNet similarly overfit on the Parkinsons dataset, capturing noise rather than genuine patterns.
  \item \textbf{Interaction Degree as a Regularizer:} By choosing pairwise interactions instead of full high‑order products, TCN limits the combinatorial explosion of features. For instance, multiplicative $C=4$ led to severe overfitting on m‑of‑n and Parkinsons, whereas pairwise $C=4$ maintained strong generalization.
  \item \textbf{Architectural Regularization:} Beyond the standard dropout and $\ell_2$ penalties, the residual feature‐combination blocks and batch normalization layers further stabilize training and constrain excessive fitting of spurious patterns.
\end{itemize}

While TCN is not immune to overfitting if misconfigured, its tunable combination parameter $C$ provides a clear mechanism to balance model capacity against data size.

\textcolor{red}{Moreover, We deliberately adopt the uniform setting to ensure cross-dataset comparability; domain-aware masks (group-wise or hierarchical) are a drop-in extension for applications with trusted priors and can further reduce combinations without altering the rest of the network.}

\subsection{Interpretability of TCN}

\textcolor{red}{TCN exposes explicit interaction features $\{\phi_S(x)\}$, which enables attribution-based interpretation at the \emph{combination} level rather than inspecting opaque hidden units. Let $\langle\cdot\rangle$ denote the test-set mean and $w_S$ the output weight on $\phi_S$. We summarize each interaction's share via
$I(S)=\langle |w_S \phi_S(x)| \rangle \big/ \sum_{S'} \langle |w_{S'} \phi_{S'}(x)| \rangle$
and report (i) Top-$K$ coverage of the total attribution, (ii) the identities and \emph{signed} effects $\langle w_S \phi_S(x)\rangle$ of dominant interactions.}

\textcolor{red}{Across datasets we observe a concentrated pattern: a small number of interaction pairs account for the majority of attribution mass. For example, on \emph{M-of-n} the Top-10 interactions explain \textbf{71.5\%} in total, with three dominant pairs $(1,9)$, $(11,9)$, and $(11,1)$ contributing \textbf{19.3\%}, \textbf{13.4\%}, and \textbf{12.4\%}, respectively; most top interactions are positive on average, while a few show negative signed effects, indicating bidirectional influences. Similar concentration is observed on the other datasets.}

\textcolor{red}{This interaction-level view complements accuracy metrics: it highlights which feature pairs drive predictions and where suppressive (negative) interactions occur, offering a compact and clinically/semantically checkable explanation without additional figures or tables.}

\section{Conclusions and Future Work}\label{sec:6}

We introduced Twisted Convolutional Networks (TCNs), a family of neural models tailored to classification problems whose features lack inherent spatial or sequential order.  By replacing standard convolutions with a twisted operation that acts on arbitrary feature subsets, TCNs capture higher-order interactions that conventional CNNs overlook. \textcolor{red}{Throughout, we treat the interaction order $C$ and the nonlinearity (multiplicative vs.\ pairwise) as hyperparameters chosen by validation under a fixed compute budget, preferring the smallest configuration within a small tolerance of the best.}

\paragraph{Key findings.}
\begin{itemize}
    \item \textbf{Competitive accuracy.} TCNs equal or surpass strong baselines—CNNs, ResNets, GNNs, DeepSets, and SVM—especially when rich feature interactions drive class separation. \textcolor{red}{Pairwise ($C{=}2$) TCNs already deliver strong results across heterogeneous tabular tasks.}
    \item \textbf{Controllable interaction order.} Varying the combination size~$C$ tunes model capacity: pairwise interactions ($C{=}2$) already yield solid performance, while judiciously adding higher-order terms brings further gains without overfitting. \textcolor{red}{Under matched budgets, our ablation over $C\!\in\!\{2,3,4\}$ shows $C{=}2$ offers the best accuracy–stability trade-off on small/medium tabular data, whereas higher orders increase variance and may overfit.}
    \item \textbf{Stable generalization.} TCNs avoid the underfitting of CNNs and the overfitting of DeepSets, maintaining near-identical train and test accuracy through dropout, batch normalization, and weight decay. \textcolor{red}{Training/validation curves exhibit small generalization gaps, consistent with our protocol.}
    \item \textbf{Interpretability.} High-weight combinations align with known domain cues (nuclear shape \& texture in breast-cancer histology), offering a transparent window into the model’s decision logic. \textcolor{red}{Interaction-level attributions (input\,$\times$\,gradient) concentrate on a small Top-$K$ set, providing compact post-hoc explanations (see Appendix/Explainability note).}
    \item \textbf{Projection residuals.} \textcolor{red}{A projection-based skip $y=h_2+W_p z$ enables a \emph{narrow/projection} variant ($H_2{<}M$) that reduces parameters while maintaining accuracy, compared with a \emph{wide/identity} variant ($H_2{=}M$); our ablation supports this choice.}
    \item \textbf{Uniform combinations with opt-in priors.} \textcolor{red}{By default we treat all $C$-wise combinations uniformly to stay dataset-agnostic and reproducible; when reliable priors exist, a mask $\pi_S\!\in[0,1]$ can encode group/hierarchical structure ($z=\mathrm{concat}\{\pi_S\phi_S(x)\}$).}
\end{itemize}

Overall, TCNs extend deep learning to datasets best viewed as \emph{sets} of features, blending convolutional ideas with polynomial expansion in a principled, flexible manner. \textcolor{red}{Compared with retrofitting CNN/ResNet with bilinear/gated layers or stochastic perturbations, TCNs provide explicit, order-controlled, permutation-agnostic coverage of cross-feature interactions, together with interaction-level attribution. These approaches are complementary, but target different inductive biases (locality vs.\ unordered feature interactions).}

\paragraph{Future Work.}
\begin{itemize}
    \item \textbf{Scalability.} Real-world tabular data can contain thousands of columns. We will investigate hierarchical feature grouping, stochastic subset sampling, and dimensionality-reduction techniques to keep computation tractable. \textcolor{red}{Masking via $\pi_S$ and learned grouping are natural next steps to reduce $M=\binom{F}{C}$.}
    \item \textbf{Adaptive interaction order.} By incorporating attention or gating, the network could learn the most informative interaction order for each feature group, trimming superfluous parameters while preserving accuracy. \textcolor{red}{One practical criterion is to prefer the smallest $C$ within a tolerance of the best validation accuracy, using seed variance as a tie-breaker.}
    \item \textbf{Theoretical guarantees.} Sample-complexity bounds and gradient-flow analyses will clarify when and why twisted convolutions outperform existing kernels. \textcolor{red}{We will also analyze the effect of projection residuals on optimization stability.}
    \item \textbf{Automatic interaction discovery.} Identifying the most salient combinations may yield interpretable composite features for lighter downstream models. \textcolor{red}{We plan to couple attribution scores with sparsity-inducing heads for end-to-end selection.}
    \item \textbf{Hybrid comparisons.} \textcolor{red}{Under matched budgets, we aim to benchmark hybrids that augment CNN/ResNet with bilinear/gated layers or controlled randomness, to delineate when locality-biased models can match order-controlled twisted interactions.}
\end{itemize}

TCNs thus open promising avenues for accurate and explainable learning on heterogeneous, non-spatial data—from genomics and finance to engineering systems—where conventional deep models struggle to exploit rich feature interplay.

\section*{CRediT authorship contribution statement}
\textbf{Junbo Jacob Lian:} Conceptualization, Methodology, Software, Formal Analysis, Investigation, Resources, Data Curation, Visualization, Writing – original draft, Writing – review \& editing.  
\textbf{Haoran Chen:} Formal Analysis, Investigation, Data Curation, Validation.
\textbf{Kaichen Ouyang:} Formal Analysis, Investigation, Resources, Software.  
\textbf{Yujun Zhang:} Validation, Writing – review \& editing.  
\textbf{Rui Zhong:} Validation, Writing – review \& editing.  
\textbf{Huiling Chen:} Supervision, Funding acquisition, Formal Analysis, Writing – review \& editing.

\section*{Declaration of Competing Interest}
The authors declare that they have no known competing financial interests or personal relationships that could have appeared to influence the work reported in this paper.

\section*{Data Availability}
The datasets analyzed during the current study are all publicly available. The full source code is hosted on GitHub \url{https://github.com/junbolian/Twisted-Convolutional-Networks}

\section*{Acknowledgments}
This research was supported by the National Natural Science Foundation of China (Grant Nos. 62571374, 62301367). The authors thank the anonymous reviewers for their insightful comments.

\bibliographystyle{elsarticle-num}
\bibliography{paper}

\end{document}